\documentclass{tlp}

\usepackage{amsmath}
\usepackage{graphicx}
\usepackage{multirow}

\usepackage{algorithm}
\usepackage{algorithmic}
\usepackage{natbib}

\usepackage{tablefootnote}

\usepackage{amssymb}
\usepackage{xspace}
\usepackage[utf8]{inputenc}
\usepackage{xcolor}  
\newtheorem{example}{Example}

\newtheorem{proposition}{Proposition}
\newtheorem{definition}{Definition}

\newcommand{\LL}{{\mathcal L}}

\newcommand{\I}{\mathcal{I}}
\newcommand{\rul}{\leftarrow}

\newcommand{\ICOAgg}[1]{\Phi^{Aggr}_{#1}}
\newcommand{\ICO}[1]{\Phi_{#1}}

\newcommand{\modelsdown}{\models_3}
\newcommand{\modelsup}{\models_3^\uparrow}

\newcommand{\HH}[2]{\mathcal{H}_{#1}#2}

\newcommand{\Tr}{\ensuremath{\mathbf{t}}}
\newcommand{\Fa}{\ensuremath{\mathbf{f}}}
\newcommand{\Un}{\ensuremath{\mathbf{u}}}

\newcommand{\ignore}[1]{}
\newcommand{\HT}{\textbf{HT}\xspace}

\newcommand{\SUM}{\mathit{SUM}}
\newcommand{\PROD}{\mathit{PROD}}

\newcommand{\KK}{\textit{KK}}
\newcommand{\lfp}{\textit{lfp}}
\newcommand{\WF}{\textit{WF}}

\newcommand{\Ult}{\textit{Ult}}
\newcommand{\ult}{\textit{ult}}
\newcommand{\glb}{\textit{glb}}
\newcommand{\lub}{\textit{lub}}
\newcommand{\Inc}{\mathbf{i}}

\newcommand{\Aggr}{\textit{Aggr}}
\newcommand{\Agg}{\textit{Agg}}
\newcommand{\ms}{\textit{ms}}

\newcommand{\GL}{\textit{GL}}
\newcommand{\FLP}{\textit{FPL}}

\newcommand{\F}{\textit{F}}
\newcommand{\TSR}{\textit{TSR}\xspace}
\newcommand{\TSRs}{\textit{TSRs}\xspace}

\newcommand{\MR}{\textit{MR}}
\newcommand{\Godel}{\textit{G3}}

\newcommand{\GZ}{\textit{GZ}}
\newcommand{\cond}{\textit{cond}}

\newcommand{\LPST}{\textit{LPST}}
\newcommand{\LB}{\textit{LB}}
\newcommand{\UB}{\textit{UB}}
\newcommand{\bnd}{\textit{bnd}}
\newcommand{\triv}{\textit{triv}}

\newcommand{\TP}[1]{T_{#1}}

\begin{document}

\lefttitle{Vanbesien, L., Bruynooghe, M., \& Denecker, M.}

\jnlPage{x}{x}
\jnlDoiYr{2022}
\doival{10.1017/xxxxx}

\title[Analyzing Semantics of Aggregate ASP Using AFT]{Analyzing Semantics of Aggregate Answer Set Programming Using Approximation Fixpoint Theory}

\begin{authgrp}
\author{\gn{Linde} \sn{Vanbesien}}
\affiliation{Department of Computer Science, KU Leuven, Belgium}
\author{\gn{Maurice} \sn{Bruynooghe}}
\affiliation{Department of Computer Science, KU Leuven, Belgium}
\author{\gn{Marc} \sn{Denecker}}
\affiliation{Department of Computer Science, KU Leuven, Belgium}
\end{authgrp}

\history{\sub{xx xx xxxx;} \rev{xx xx xxxx;} \acc{xx xx xxxx}}

\maketitle

\begin{abstract}
Aggregates provide a concise way to express complex knowledge. The problem of selecting an appropriate formalisation of aggregates for answer set programming (ASP) remains unsettled. This paper revisits it from the viewpoint of Approximation Fixpoint Theory (AFT). We introduce an AFT formalisation equivalent with the Gelfond-Lifschitz reduct for basic ASP programs and we extend it to handle aggregates. We analyse how existing approaches relate to our framework. We hope this work sheds some new light on the issue of a proper formalisation of aggregates. This paper is under consideration for acceptance in TPLP.
\end{abstract}


\section{Introduction}
Aggregate expressions are very useful and have been added to classical logic, query languages, constraint languages, and also to logic programming (LP) and answer set programming (ASP).
The effort it takes to add aggregates to (syntax and semantics of) a logic is very
language dependent. E.g., to extend first order logic (FO) with a
\emph{Count} aggregate, we extend the definition  of ``term'' with a new inductive rule: 
``{\em If $a_1,\ldots,a_n$ are variables, and $\psi$ a formula, then
$Count(\{(a_1,\ldots,a_n),\psi\})$ is a term}''
and the definition of ``interpretation of a term $t$ in structure $I$'' (used in the definition of $\models$) with: 
``{\em If $t = Count(\{(a_1,..,a_n),\psi\})$ then $t^I$  is $\# \{(d_1,\ldots,d_n)\in Dom(I)^n | I[a_1:d_1,\ldots,a_n:d_n] \models \psi$\}, i.e., the number of tuples that satisfy $\psi$ in $I$.}''. The method is simple and follows Frege's compositionality principle. 

In  LP and ASP, it is much more difficult. Research into extensions  with aggregates  (well-founded and stable semantics)  started with the work of \cite{KempStuckey91}.  Many approaches exist, but so far no consensus on how to handle aggregates in those logics has been reached.

We propose a framework for defining semantics for extensions of LP and ASP based on Approximation Fixpoint Theory (AFT) introduced by \cite{denecker2000approximations} and \cite{denecker2004ultimate}. AFT is an abstract lattice theoretic formalization of constructive methods for non-monotonic operators. It defines different types of constructions and fixpoints to a lattice operator in an approximation space, including supported, Kripke-Kleene (KK), stable and well-founded (WF) fixpoints. The theory has been applied to a range of non-monotonic logics to characterize existing as well as new semantics: e.g., LP and ASP as shown in the paper by \cite{denecker2012approximation}, autoepistemic and default logic as shown in the paper by\cite{DeneckerMT03}, higher order LP as shown in the paper by \cite{CharalambidisRS18}, argumentation frameworks and abstract dialectal frameworks as shown in the papers by \cite{Strass13} and \cite{bogaerts2019weighted}. AFT has been applied to aggregate LP and ASP resulting in the ultimate stable and well-founded semantics in the work by \cite{DeneckerPB01} and the broader framework in the paper by \cite{pelov2007well} where  stable and well-founded semantics are induced by a choice of a 3-valued truth function. 

Here, we clarify and expand this work. First, to make the AFT framework more accessible to the ASP community, we show how each approximation truth assignment can be broken up in a lower and an upper ternary satisfaction relation which, in the context of ASP can be easily related to the reduct approach  originally used by \cite{GL-reduct} to define stable semantics. Then we focus on aggregate ASP using examples from the literature. Where possible, we consider aggregate atoms with positive and negative literals as conditions. However, the semantics described by \cite{Gelfond2019vicious} does not allow \textit{negation by default} inside an aggregate atom, therefore we only consider positive conditions for this specific case. It is shown that not only the semantics of \cite{DeneckerPB01,pelov2007well} but also  those of \cite{liu2010logic}  and \cite{Gelfond2019vicious} are instances of our framework. But not all proposed semantics for aggregate ASP semantics belong to our framework; for example those of \cite{ferraris2011logic,marek2004set,faber2011semantics}. We investigate the reason for this. The paper contributes to the discussion about semantics for Aggregate ASP by clarifying some important principles of NMR and by showing where they are applied and where other principles are applied. 

\section{Approximation Fixpoint Theory}\label{sec:AFT}
Here, we recall the basics of AFT from the work by \cite{denecker2000approximations}.
In many non-monotonic languages, a theory defines a semantic lattice\footnote{A lattice $\langle L,\leq \rangle$ is a partially ordered set where each subset $S$ has a least upper bound $\lub(S)$ and a greatest lower bound $\glb(S)$.} operator $O: L\rightarrow L$. If $O$ is monotone,  its least fixpoint is often taken as the semantics of the theory. Otherwise, AFT can be applied. The first step is to associate the approximation space $L^2$ to $L$. A pair $(x,y)\in L^2$ is an approximation of any $z\in[x,y]$. With $x \leq y$, the interval is non-empty and the pair is {\em consistent}. $L^c$ is the subspace of consistent pairs. $L^2$ and $L^c$ possess (i) a precision order, $(x,y)\leq_p (u,v)$ if $x\leq u$ and $y\geq v$ and (ii) the embedding of $L$, namely the set of {\em exact pairs} $(x,x)$ which approximate only $x$. The least precise point is $(\bot,\top)$, with $\bot=glb(L),\top=lub(L)$.

The second step is to assign an approximating operator $A$ to $O$: a $\leq_p$-monotone operator on $L^2$ or $L^c$ such that if  $(x,y)$ approximates $z$ then $A(x,y)$ approximates $O(z)$; an approximator on $L^2$ also has to be symmetric:  $A(x,y)=(u,v)$ iff $A(y,x)=(v,u)$. Thus, increasing the precision of the input to an approximator $A$ increases the precision of the output. With an approximator $A$, several types of fixpoints are definable. The least fixpoint (lfp) construction $(\bot,\top)$, $A(\bot,\top)$, \dots, $A^\alpha(\bot,\top)$, \dots produces the  KK fixpoint $\KK(A)=\lfp(A)$. The KK fixpoint is a pair $(x,y) \in L^2$. If exact, $x (= y)$ is the only fixpoint of $O$. Otherwise, it approximates all fixpoints of $O$, including ``self-supported'' ones that are often not minimal in L. Stable and WF fixpoint definitions contain mechanisms  to reduce self-support.\footnote{The intuition of  self-support is not easily explained in an algebraic setting but shows intuitively in the logic program $\{p \rul p.\ \  q \rul \neg p\}$. It has two minimal fixpoints $\{p\}$ and $\{q\}$, but $\{p\}$ is self-supported (in $p$) while $\{q\}$ is not. The second is the unique stable and well-founded fixpoint. } A stable fixpoint $x\in L$ is one such that $x=\lfp(\lambda z: A(z,x)_1)$, where $A(z,x)_1$ is the first component of the pair $A(z,x)$. The WF fixpoint $\WF(A)$ is the least precise pair $(x,y)$ with $x=\lfp(\lambda z: A(z,y)_1)$ and $y=\lfp(\lambda z: A(x,z)_2)$.  It is the least precise fixpoint of the monotone operator $(x,y)\mapsto (\lfp(\lambda z:A(z,y)_1),\lfp(\lambda z:A(x,z)_2))$. We have that $\KK(A)$ and $\WF(A)$ are consistent and for each stable fixpoint $x$, $\KK(A)\leq_p \WF(A)\leq_p x$ and  $x$ is a minimal fixpoint of $O$.\footnote{In case of an $L^c$-operator,  the domain of $\lambda z:A(z,y)_1$  is $[\bot,y]$ and that of $\lambda z: A(x,z)_2$ is $[x,\top]$  while the range of both operators is $L$. So, it is possible  that the iterated lfp  construction of one of these operators terminates in a point outside the operator domain in which case the operator has neither a fixpoint nor a lfp. To accommodate, we call $x$ a stable fixpoint of $A$ if the lfp of $\lambda z:A(z,y)_1$ exists and is equal to $x$.  In the paper by \cite{denecker2004ultimate}, it was proven that every $L^c$-approximator $A$ is expandable to $L^2$ approximators, and that each such an expansion has the same KK, WF and stable models as $A$. Therefore, we spend little attention to the difference between $L^2$ and $L^c$. }

AFT induces relationships between fixpoints of different approximators of $O$. If approximator $A$ is pointwise less precise than $B$, then $\KK(A)\leq_p \KK(B)$, $\WF(A)\leq_p  \WF(B)$ and any stable fixpoint of $A$ is a stable fixpoint of $B$. Thus, with increasing precision of the approximator,  KK and WF fixpoints increase in precision, and the set of stable fixpoints grows. There exists  a most precise approximator $\Ult_O$ of $O$, with the most precise KK and WF fixpoint and the largest set of stable fixpoints. For consistent pairs $(x,y)$, $\Ult_O(x,y)$ is the most precise pair approximating $O([x,y])$, i.e.,  $(\glb(O([x,y]),\lub(O([x,y])))$. It follows, perhaps surprisingly, that if an even moderately precise approximator $A$ has a stable fixpoint $x$,  then $x$ is  approximated by the most precise WF fixpoint $\WF(\Ult_O)$ associated with $O$ (but keep in mind that unprecise  approximators are unlikely to have stable fixpoints). This counter-intuitive fact tells us that in all cases, if stable fixpoints exist, they are in the proximity of the most precise WF fixpoint that can be associated to $O$. This may explain the good quality of stable semantics in capturing intuitions. But while KK and WF are constructive, stable semantics is not really. It only has a  constructive test: testing if $x$ is stable is by testing if $x$ is the limit of the $\lfp$ construction of $\lambda z: A(z,x)_1$. 

We now sketch how to use AFT to define constructive semantics for programs based on some logic $\LL$.\footnote{For ease of discussion, this work only considers Herbrand interpretations and ground programs.}
We assume  $\LL$ is a first-order logic with a (Herbrand) model semantics defined using a 2-valued truth function $\HH{}^2$, or equivalently, a satisfaction relation $\models_2$ (where $I\models_2 \phi$ iff $\HH{I}^2(\phi)=\Tr$). An $\LL$-program is then defined as follows: 

\begin{definition}[$\LL$-program]
An $\LL$-program is a set of rules $r$ of the form $p \leftarrow \psi$ such that the body $\psi$ is a formula of $\LL$ and the head $p$ is a propositional atom. 
\end{definition}

Importantly, $\rul$ is not a connective of $\LL$ but AFT defines its meaning as a {\em construction operator}. LP and (non-disjunctive) ASP are instances of this where $\LL$ is simply the logic of conjunctions of literals $p$ or $\neg p$ under standard interpretation.
Given an $\LL$-program $P$, the corresponding lattice $L,\leq$ is the set of $P$'s Herbrand interpretations ordered by the truth order ($\Fa<\Tr$). The sets $L^c$ and $L^2$ correspond to 3- and 4-valued interpretations. Any 3- or 4-valued interpretation $\I$ can be split into a pair $(I,J)$ by  splitting truth values as in $\Tr\sim(\Tr,\Tr), \Fa\sim(\Fa,\Fa), \Un\sim(\Fa,\Tr)$ and $\Inc\sim(\Tr,\Fa)$. If $\I$ is 3-valued, then $I\leq J$, and $I$ is a lower bound,  $J$ an upper bound of $\I$. The 3- and 4-valued structures are equipped with a truth order, which isomorphically corresponds to the product order $\leq$ of $L^2$ and $L^c$, and with a precision order $\leq_p$ ($\Un <_p \Tr <_p \Inc, \Un<_p\Fa<_p\Inc$ which corresponds to the precision order of $L^2$ and $L^c$. Any truth or precision monotone operator $\Gamma$ on 2-, 3- or 4-valued structures corresponds to a truth or precision monotone operator on $L, L^c, L^2$. 

An $\LL$-program $P$ is characterised by the immediate consequence operator $\TP{P}:L\rightarrow L$. This operator fulfills the role of $O$, the approximated operator. The immediate consequence operator $\TP{P}:L\rightarrow L$ for a program $P$ is such that $\TP{P}(I)=J$ if for every  ground atom $p$, $\HH{J}^2(p) = \lub_\leq(\{\HH{I}^2(\psi) | (p \leftarrow \psi) \in P\})$.  There are two ways to define AFT semantics using an approximator $A (= A_P)$ for $\TP{P}$. 

One way is to use $A_P = \Ult(\TP{P})$, the most precise approximator of $\TP{P}$ in $L^c$ leading to ultimate versions of the family of AFT semantics as described by \cite{denecker2004ultimate} and used by \cite{DeneckerPB01} for defining semantics of Aggregate LP. This is the most precise approach, but computationally costly.  The other way is to extend $\LL$'s truth assignment to 3- or 4-valued interpretations and by  interpreting $\TP{P}$'s definition in this broader context. E.g., a 3-valued truth assignment $\HH{}^3$ induces a 3-valued immediate consequence operator $\ICO{P}$ where $\ICO{P}(\I)=\I'$ if for every atom $p$, $\HH{\I'}^3(p) = \lub_{\leq}\{\HH{\I}^3(\psi)|(p\leftarrow \psi) \in P\}$. 
The  operator $\ICO{P}$ corresponds isomorphically to an $L^c$ approximator $A_P$ on pairs $I<J$ of 2-valued interpretations.  But  for $A_P$ to be an approximator of $\TP{P}$, the 3-valued truth assignment $\HH{}^3$ should satisfy a condition introduced for 3-valued logic by \cite{Kleene52}:

\begin{definition}[Regular truth assignment] 
A 3-valued truth assignment $\HH{}^3$ of $\mathcal{L}$ is  {\em regular} iff for all formulas $\psi$, for all 3-valued structures $\I$ interpretating $\psi$: (1) (extension of $\HH{}^2)$ if $\I$ is 2-valued, then  $\HH{\I}^2(\psi)=\HH{\I}^3(\psi)$ and  (ii) (precision monotonocity) if $\I\leq_p \I'$ then $\HH{\I}^3(\psi) \leq_p \HH{\I'}^3(\psi)$ .\footnote{For the 4-valued case, an additional condition is symmetry.} 
\end{definition}

E.g., Kleene's strong 3-valued truth assignment $\HH{}^{SK}$ of FO (introduced by \cite{Kleene52}) and Belnap's 4-valued extension (introduced by \cite{belnap1977useful}) are regular. They induce multi-valued extensions $\ICO{P}$ of  $\TP{P}$ first introduced by \cite{fitting1985kripke}. Later, $\ICO{P}$  was  found to  correspond to an approximator $A_P$ of $\TP{P}$ whose KK, WF and stable fixpoints correspond to the semantics of the same name. 

In LP and ASP, it is often taken for granted that LP's nonmonotonicity is due to  its non-classical negation $not$. But it is evident in AFT-based semantics, that the main non-classical connective is the rule operator: its semantics is defined via operators and constructive processes while negation in bodies is treated like the other FO connectives, using three-valued logic only for approximation of the standard classical connectives.  

The AFT road  is quite unlike other semantic techniques in ASP.  In the next section, we  reformulate the  framework in more accessible terms for the ASP community. 

\section{Ternary Satisfaction Relations}

Ternary satisfaction relations were used originally by \cite{liu2010logic} in the context of Aggregate ASP semantics where they were called \textit{sub-satisfiability relations}. 

We still assume  a base logic $\LL$ equipped with  satisfaction relation $\models_2$. Below, we restrict ourselves to  3-valued interpretations corresponding to pairs $(I,J)$ of 2-valued  Herbrand interpretations  $I\subseteq J$ (but extension to non-Herbrand interpretations is possible).

\begin{definition}[Ternary satisfaction relations]
A \emph{ternary satisfaction relation} (\TSR) $\models_3$ of  $\mathcal{L}$ is a relation between pairs $(I,J)$ of  interpretations such that $I\subseteq J$, and formulas $\psi$ of $\mathcal{L}$ such that  $I\models_2 \psi$ iff  $(I,I)\models_3 \psi$.  It is \emph{lower-monotone}  if $(I, J) \models_3 \psi$ implies  $(I', J) \models_3 \psi$ when $I\subseteq I' \subseteq J$. It is  \emph{lower-regular} (\emph{upper-regular}) if $(I, J) \models_3 \psi$ implies $(I', J') \models_3 \psi$ when  $(I, J) \leq_p (I', J')$  ($(I', J') \leq_p (I, J)$). 
\end{definition}

For any pair of a lower-regular \TSR   $\modelsdown$  and an upper-regular \TSR $\modelsup$, it always holds that $\modelsdown\subseteq\modelsup$ since $(I,J)\modelsdown\psi$ implies $(I,I)\models_2\psi$ which implies $(I, J)\modelsup \psi$. Also, a three-valued truth-function $\HH{}^3$  corresponds one to one to pairs $(\modelsdown,\modelsup)$ of \TSRs satisfying  $\modelsdown \subseteq \modelsup$. The correspondence is: 
(i) $(I, J) \modelsdown \psi$ iff $\HH{(I, J)}^3(\psi) = \Tr$ and
(ii) $(I, J) \modelsup \psi$ iff $\HH{(I, J)}^3(\psi) \in \{\Tr, \Un\}$.

\begin{proposition}
\label{prop:derived3truth}
$\HH{}^3$ is regular iff $\modelsdown$ is a lower- and $\modelsup$ an upper-regular \TSR.\footnote{All proofs are in the supplementary material corresponding to this paper at the TPLP archives.}
\end{proposition}

Taking $\LL$ as FO and $\HH{}^3$ as the strong Kleene truth assignment, it is a folk result\ignore{Feferman et al hebben dat wellicht voor het eerst bewezen, de PhD van Johan Wittocx  bevat dit resultaat: zoek feferman erin} that $(I,J)\modelsdown \psi$ if $\psi$ evaluates to true when interpreting  all positively occurring atoms  in $I$ and all negatively occurring ones in $J$. 
For  $(I,J)\modelsup \psi$, exchange the roles of $I$ and $J$. 

For each \ignore{non-disjunctive} $\LL$-program $P$, the lower- and upper-regular \TSR induce two distinct operators on consistent pairs $I\subseteq J$:  $A_P^{\modelsdown}(I, J) = \{p| \exists (p \leftarrow \psi) \in P: (I, J)\modelsdown \psi\}$ and $A_P^{\modelsup}(I, J) = \{p| \exists  (p \leftarrow \psi) \in P: (I, J)\modelsup \psi \}$. Now, we define $A_P(I, J) = (A_P^{\modelsdown}(I, J), A_P^{\modelsup}(I, J))$.

\begin{proposition} If  $\modelsdown$ and $\modelsup$ are lower- and upper-regular \TSRs, then $A_P$ is an $L^c$ approximator. Moreover it is isomorphic to the 3-valued $\ICO{P}$ induced by the 3-valued truth assignment $\HH{}^3$ combining $\modelsdown$ and $\modelsup$.
\end{proposition}

Now, supported, KK, WF and stable models of the \ignore{non-disjunctive} $\LL$-programs $P$ can be defined in terms of  $\modelsdown$ and $\modelsup$.
Interestingly, $J$ is a stable model of $P$ iff $J$ is the least fixpoint of $A_P^{\modelsdown}(I,J) =\lambda I\in [\bot,J]: \{p| \exists (p \leftarrow \psi) \in P: (I, J)\modelsdown \psi\}$. No need of $\modelsup$! Clearly there exists an asymmetry between truth and falsity in stable models. While information about the truth of formulas, encoded by $\modelsdown$, is essential to determine the stable fixpoints of a program, information about their falsity, given by $\modelsup$, is disregarded.

\section{Generalizing the concept of answer set}\label{sec:ASP}
Here, we generalize stable models of \ignore{(disjunctive)} $\LL$-programs to answer sets. Now, $\LL$, the logic of the rule bodies, has, besides a  satisfaction relation $\models_2$ also a \TSR $\modelsdown$. 

\begin{definition} \label{def:GAS} $I$ is an answer set of $P$ if (1) for every $(p\leftarrow \psi)\in P$, if $I\models_2\psi$ then $I\models_2 p$; (2) there is no $J\subset I$ such that for every $(p\leftarrow \psi)\in P$, if $(J,I)\modelsdown \psi$ then $(J,I)\modelsdown p$.
\end{definition}

\begin{proposition}[semi-constructive answer sets]\label{prop:semi-constructive}
If $\modelsdown$ is lower-monotone, then for \ignore{non-disjunctive  }$\LL$-programs $P$, $I$ is an answer set  of $P$ iff $I$ is the  limit of the increasing sequence $\langle I_\alpha\rangle_{\alpha\geq 0}$ where (1) $I_0=\emptyset$, (2) $I_{\alpha+1}=A_P^{\modelsdown}(I_\alpha,I)$ if $I_\alpha\subseteq I$, (3) $I_\lambda = \bigcup_{\alpha<\lambda}I_\alpha$ for limit ordinal $\lambda$. 
\end{proposition}

In general, the (transfinite) fixpoint sequence may leave $[\emptyset,J]$,  or end up with a fixpoint $I \subseteq J$. In case it is $J$, it is an answer set. 
Let $\modelsdown$ be lower-regular. Combining it with {\em any} upper-regular \TSR $\modelsup$ induces a regular truth assignment $\HH{}^3$, as well as an entire family of AFT fixpoints and models of \ignore{non-disjunctive} $\LL$-programs: KK-models, WF-models and AFT-stable models approximated by the WF-models. The AFT-stable models in this framework depend only on $\modelsdown$ and they correspond exactly to answer sets of $\modelsdown$, since a lower-regular \TSR is also lower-monotone. 

\begin{proposition}
Answer sets and AFT-stable models coincide for $\LL$ programs with a lower regular \TSR.
\end{proposition}

\label{sec:GL}
\label{GL-satisfaction}

To finish this section, we analyze the link with the original definition of answer set given by \cite{GL-reduct}. There,  $\LL$ is the logic of ground sets/conjunctions of literals with FO's standard satisfaction relation $\models_2$. Let $P$ be an \ignore{disjunctive } $\LL$-program. 
\begin{definition}[Gelfond-Lifschitz reduct and answer set]
The Gelfond-Lifschitz reduct (defined by \cite{GL-reduct}) $P^J$ of $P$ for an interpretation $J$ is obtained from $P$ by deleting 
\begin{itemize}
\setlength{\itemindent}{1em}
    \item all rules with a negative literal $\neg l$ such that $l \in J$.
    \item all negative literals in the bodies of the remaining rules.
\end{itemize}
$J$ is a GL-answer set of $P$ if $J \models_2 P^J$ and there is no $I\subset J$ such that $I\models_2 P^J$. 
\end{definition}

We now identify the lower-regular \TSR $\models_{GL}$ such that answer sets of \ignore{disjunctive }programs induced by $\models_{GL}$, coincide with GL-answer sets. The \TSR that is needed evaluates bodies $\psi$ in pairs $(I,J)$ by interpreting atomic literals of $\psi$ in $I$ and negative literals in $J$. But as explained in the previous section, that is how the lower-regular \TSR $\models_{SK}$ of the strong Kleene truth assignment operates: $(I,J)\models_{SK} \psi$ iff atoms in $\psi$ hold in $I$ and negative literals hold in $J$. Thus, $\models_{GL}$ is the restriction of $\models_{SK}$ to conjunctions of literals. With this in mind, the following proposition is straightforward.  

\begin{proposition} For $I\in[\bot, J]$, $I\models_2 P^J$  iff  for every rule $p\leftarrow\psi\in P$, if $(I,J)\models_{GL} \psi$ then $(I,J)\models_{GL} p$. \ignore{If $P$ is non-disjunctive, then}$J$ is a GL-answer set of $P$ iff $J$ is an AFT-stable model of $P$ under strong Kleene truth assignment. 
\end{proposition}
Thus, this type of answer set fits in the AFT-landscape of KK, WF and stable models. 

\section{Aggregates Programs in the AFT Framework}
\paragraph{Aggregate Programs}\label{sec:aggr}
For the remainder of the text we will consider $\LL$-programs where $\LL$ is the logic of conjunctions of literals and positive aggregate atoms.\footnote{In many semantics, including AFT semantics, a negated aggregate atom can always be represented by its dual positive aggregate atom. E.g., $\neg(SUM(\{1 :s\})>0)$ corresponds to $SUM(\{1 :s\})\leq 0$. }

An aggregate atom $a^\Aggr$ is of the form: $\Agg(\{a_1: \cond_1,..., a_n: \cond_n\})*w$ with aggregate symbol $\Agg$ (e.g., $SUM$), comparison connective $*$ (e.g., $\leq, =, \neq,\dots$), numerical value $w$ and multiset $\{a_1: \cond_1,..., a_n: \cond_n\}$ wher each $cond_i$ is a literal and each $a_i$ is a weight. The $\models_2$ relation for $\LL$ is naturally extended with a rule for evaluating aggregate atoms, so also $\TP{P}$ is defined. This enables the first approach to apply AFT on this type of programs, using the 3-valued ultimate approximator $\Ult_{\TP{P}}$ yielding  the ultimate KK, WF and stable semantics. Up to the syntax, this is the semantics of \cite{DeneckerPB01}. The second approach to apply AFT is based on defining a regular 3-valued truth assignment for aggregate atoms, leading to a three valued operator $\ICOAgg{P}$. Up to the syntax, this was the approach followed by \cite{pelov2007well}.

We now start the study of existing approaches for handling aggregates in ASP. Some of these use more extensive programs than the ones analysed here, however our analysis only considers the simpler $\LL$-programs. Proposition \ref{prop:semi-constructive} shows that there is a constructive test for answer sets of non-disjunctive $\LL$-programs for semantics with lower-monotone \TSR's. A lower-regular \TSR is lower-monotone by definition. Thus, this constructive test is applicable for all semantics that fit in the AFT-framework. However, not all semantics for ASP-programs in the literature have lower-regular or even lower-monotone ternary satisfaction relations. 

Before we start, we define the precision order on the \TSRs analogous to the precision order on truth assignments defined by \cite{pelov2007well}. 

\begin{definition}[Precision relation over lower ternary satisfaction relations]
A \TSR $\models_a$ is less precise than a \TSR $\models_b$, or $\models_a \leq_p \models_b$, iff for every formula $\psi$ and every pair of two-valued interpretations $(I, J)$: $(I, J) \models_a \psi$ implies $(I, J) \models_b \psi$.
\end{definition}

\begin{proposition}\label{prop:precisionmonotone}
Let $\models_a, \models_b$ be \textit{TSR}s that coincide with $\models_{GL}$ on aggregate free bodies. If $\models_a \leq_p \models_b$ and $J$ is an answer set associated with $\models_a$ (an $a$-answer set), then $J$ is a $b$-answer set. 
\end{proposition}

\subsection{Approaches that fit in the AFT framework}

\paragraph{\cite{pelov2007well}}
This paper introduces several regular truth assignments for aggregate atoms. This is equivalent with expanding the lower- and upper-regular \TSR  with a rule for aggregate atoms. For defining answer sets for the syntax of this paper, the lower one suffices.\footnote{The formalism of \cite{pelov2007well} is much richer including aggregate atoms under negation, and its semantics requires lower- and upper-regular \textit{TSR}s defined inductively in terms of each other. } The least precise approximation, $\HH{}^{triv}$ assigns to an aggregate atom $a^{Aggr}$ the same value as $I$ and $J$ when $I$ and $J$ agree on the conditions in $a^{\Aggr}$ and $\Un$ when they disagree. The corresponding lower \TSR is:

\begin{definition}[$\models_{\triv}$] \label{def:triv} 
$\models_{\triv}$ extends $\models_{\GL}$ with: $(I, J) \models_{\triv} a^\Aggr $ iff $J\models_2 a^{Aggr}$ and ${cond_i}^I={cond_i}^J$ for every condition $cond_i$ in $a^\Aggr$.
\end{definition}

The most precise regular truth assignment  $\HH{}^{\ult}$ assigns $\Tr$ ($\Fa$)  to an aggregate expression in $(I,J)$ if it is $\Tr$ ($\Fa$) in every $Z\in [I,J]$. Otherwise, it assigns $\Un$. The corresponding lower \TSR is:

\begin{definition}[$\models_{ult}$] \label{def:ult}
$\models_{\ult}$ extends $\models_{\GL}$ with: $(I, J) \models_{\ult} a^{\Aggr} $ iff for each $Z$ such that $I \subseteq Z \subseteq J$: $Z \models_2 a^{\Aggr}$.
\end{definition}

\cite{pelov2004semantics} shows that for stratified aggregate programs (where predicates in aggregate expressions are defined at a lower level), the trivial and the ultimate truth assignments lead to the same semantics. 

While very precise, \cite{pelov2004semantics} shows that the complexity of computing KK, WF and stable models under $\models_{\ult}$ moves to the next level of the polynomial hierarchy. To avoid this, he offers a less precise alternative called  the \emph{bounded} truth assignment. 

Phrased in terms of the present aggregate programs, it uses functions $\LB_{\Agg}, \UB_{\Agg}: \mathcal{P}(\Tilde{D_1}) \rightarrow D_2$ that maps any three-valued multiset $\{\ms\}^{\I}$ to respectively the minimum and the maximum of $\{Agg(\{ms\})^{I'}|I'\in [I, J]\}$; i.e., $\LB_{\Agg}$ represents the lower bound for the aggregate function $\Agg$ on the possible multisets and $\UB_{\Agg}$ the upperbound.
The truth value for aggregate atoms with sum and product is based on these bounds. The corresponding lower \TSR is:

\begin{definition}[$\models_{\bnd}$] \label{def:bnd}
$\models_{\bnd}$ agrees with $\models_{\ult}$ except for aggregate atoms of the form $\Agg(\{\ms\})*w$ with $\Agg \in \{\SUM, \PROD\}$ and $* \in \{=, \neq\}$ 
\begin{itemize}
\setlength{\itemindent}{2em}
    \item $(I, J) \models_{\bnd} \Agg(\{\ms\})=w $ iff $\LB_{\Agg}(\{\ms\}^{(I, J)}) = w = \UB_{\Agg}(\{\ms\}^{(I, J)})$.
    \item $(I, J) \models_{\bnd} \Agg(\{\ms\}) \neq w $ iff $\LB_{\Agg}(\{\ms\}^{(I, J)}) > w$ or $\UB_{\Agg}(\{\ms\}^{(I, J)}) < w$.
\end{itemize}
\end{definition}

\cite{pelov2004semantics} lists polynomial algorithms to compute both bounds for all aggregate atoms discussed in his thesis. The same holds for the common aggregate atoms in ASP.  Consequently, the complexity of computing the different types of models remains on the same level as for the non-aggregate case. Yet, the bound semantics is precise enough to solve many useful aggregate programs with recursion over the aggregates. 

\begin{proposition}
The \TSRs $\models_{\triv}$, $\models_{\ult}$ and $\models_{\bnd}$ are lower-regular. Since $\models_{triv}\leq_p \models_{bnd} \leq_p \models_{\ult}$, an answer set of $\models_{triv}$ is one of $\models_{bnd}$, and one of $\models_{bnd}$   is one of $\models_{\ult}$.
\end{proposition}

\paragraph{ Liu,  Pontelli, Son and Truszczy{\'n}ski (2010).}
They introduced a kind of \TSR to  define semantics for abstract constraints. While the restrictions imposed on these relations are different and do not necessarily fit into AFT, their main example, the sub-satisfiability relation as proposed by \cite{son2007answer} does. For any abstract constraint $\alpha$: $(I, J) \models_{\LPST} \alpha$ if and only if for each interpretation $Z$ such that $I \subseteq Z \subseteq J$, it holds that $Z \models_2 \alpha$. 

\begin{definition}[$\models_{\LPST}$]
The \TSR $\models_{\LPST}$ extends $\models_{\GL}$ with:
If $a^{\Aggr}$ is an aggregate atom, then $(I, J) \models_{\LPST} a^{\Aggr} $ iff for each $Z$ such that $I \subseteq Z \subseteq J$: $Z \models_2 a^{\Aggr}$.
\end{definition}

This is the same satisfaction relation as $\models_{\ult}$ in Definition~\ref{def:ult}, hence it is lower-regular and defines the same answer sets.

\paragraph{\cite{Gelfond2019vicious}.}
They construct a reduct for aggregate programs with respect to a three-valued interpretation. We only consider the case where the interpretation is two-valued. \cite{Gelfond2019vicious} allow two kinds of negation:   \textit{negation by default},  which corresponds to negation as presented in this paper, and \textit{explicit negation}, which is not a part of the syntax of the programs considered here but can be simulated by the well-known translation of  explicitly negated atoms of a predicate $p$ into atoms of a newly introduced predicate $p^*$. Since \cite{Gelfond2019vicious} do not allow default negation within an aggregate atom, here it suffices to consider programs with positive conditions inside an aggregate atom. The reduction process is split into two main parts. The first part constructs a reduct regarding the aggregate atoms. It consists of two steps:
\begin{enumerate}
    \item Removing all rules with aggregate atoms that evaluate to $\Fa$ in the interpretation.
    \item Replacing every remaining aggregate atom by the conjunction of the subset of its conditions that are $\Tr$ in the interpretation. 
\end{enumerate}

In other words, given a rule $r$ in a program $P$: $p \leftarrow l_1 \wedge ... \wedge l_n$, such that for an $l_i$ it holds that $l_i = Agg(\{a_1: \cond_1,..., a_n: \cond_n\}) * w$, then the rule is deleted in the reduct $P^J$ if $l_i$ evaluates to $\Fa$ in $J$. Otherwise the rule is replaced by $p \leftarrow l_1 \wedge ... \wedge l_{i-1} \wedge l_{i+1} \wedge ... \wedge l_n \wedge (\bigwedge \{\cond_j \in \{ \cond_1,\ldots,\cond_n\}| J \models_2 \cond_j\}$. 

The second part transforms the preliminary reduct after the first phase to its Gelfond-Lifschitz reduct. In this way, it preserves the capability to deal with ordinary propositional atoms. From this reduct one can inductively define the \TSR $\models_{\GZ}$:

\begin{definition}[$\models_{\GZ}$]
$\models_{\GZ}$ extends $\models_{\GL}$ with:
Let $a^{\Aggr} = \Agg(\{a_1: \cond_1,..., a_n: \cond_n\}) * w$. $(I, J) \models_{\GZ} a^{\Aggr} $ iff $J\models_2 a^{Aggr}$ and $(I, J) \models_{\GZ} \bigwedge \{\cond_j \in \{ \cond_1,\ldots,\cond_n\}| J \models_2 \cond_j\}$.
\end{definition}

\begin{proposition}
For aggregate programs containing only positive conditions in aggregate atoms, the \TSR $\models_{\GZ}$ is identical to the \TSR $\models_{\triv}$ and lower-regular for consistent pairs, i.e., with $(I, J)$ a consistent pair,
$(I, J) \models_{\GZ} a^{\Aggr} $ iff $(I, J) \models_{\triv} a^{\Aggr}$. 
\end{proposition}

\paragraph{Precision Complexity Trade-off.}
One expects more effort gives more precise approximations. From theorem 7.4 in the paper by \cite{pelov2007well}, it follows that if the evaluation of an expression with respect to a lower-regular ternary satisfaction relation $\modelsdown$ is polynomially computable, then checking whether or not a model is an answer set is in $P$ and deciding whether an answer set for a program exists, is in $NP$.  
This is the case for $\models_{triv}$ and $\models_{bnd}$ and for instances of the framework that coincide with $\models_{triv}$; for instances that coincide with $\models_{ult}$ the check problem is in $NP$ and the exists-problem is in $\Sigma_2^P$. 

\subsection{Other Ternary Satisfaction Relations}\label{sec:Other}
In the AFT-framework, semantics of ASP aggregate atoms are based on lower regular \TSRs. As we show in this section, also other well-known semantics can be characterized using \TSRs, however, they are not regular. This is no coincidence. The definition of answer set semantics in terms of \TSRs 
strongly resembles another well-known semantic method of ASP, namely using the logic of here-and-there (\HT) (for an overview of \HT and its applications to ASP, see the work by \cite{cabalar2017stable}). Due to very different points of view on answer sets, the two frameworks obtain different requirements for the \TSRs. AFT treats answer sets as the result of constructive processes; the rule operator serves to produce them. A production is safe if the \TSR is lower-regular. In contrast, \HT takes a non-constructive take on answer sets. In \HT, extensions build on the inherently non-regular \TSRs derived from the three-valued logic \Godel \xspace introduced by \cite{godel1932intuitionistischen} where the rule operator is treated as \HT-material implication.

\paragraph{\cite{marek2004set}.}
They study Set Constraints (SC) Programming. It builds an \textit{NSS-reduct} for an SC-program. \cite{liu2010logic} prove that this semantics for set constraints is also obtained by the following satisfaction rule for a set constraint $\alpha$: $(I, J)\models_{\MR} \alpha$, if $J\models_2 \alpha$ and there exists an interpretation $Z \subseteq I$ such that $Z\models_2 \alpha$. This leads to the \TSR $\models_{\MR}$:

\begin{definition}[$\models_{\MR}$]
$\models_{\MR}$ extends $\models_{\GL}$ with:
$(I, J) \models_{\MR} a^{\Aggr} $ iff $J\models_2 a^{\Aggr}$ and there exists an interpretation $Z$ such that $Z \subseteq I$ and $Z \models_2 a^{\Aggr}$.
\end{definition}

Consider the aggregate atom $\SUM(\{1:p, -1:q\}) \geq 0$\ and two intervals, namely $(\emptyset, \{p, q, s\}) <_p  (\emptyset, \{q\})$.
We have that $(\emptyset, \{p, q, s\}) \models_{\MR} \SUM(\{1:p, -1:q\}) \geq 0$ while $(\{p\}, \{q\}) \not\models_{\MR} \SUM(\{1:p, -1:q\}) \geq 0$. Hence, $\models_\MR$ is not lower-regular.

\begin{proposition}
(i) $\models_{\MR}$  extends $\models_2$, i.e., $(I,I) \models_{\MR} \psi $ iff $I \models_2 \psi$. 
(ii) $\models_{\MR}$ is 
lower-monotone, i.e., if $I \subseteq I'$ and $(I, J) \models_{\MR} \psi$, then $(I', J)\models_{\MR} \psi$. 
\end{proposition}

While non-regular, the ternary relation $\models_{\MR}$ still extends the satisfaction relation $\models_2$ and induces a monotone operator $\lambda I: A_{\models_{MR}}(I,J)$. Thus, an answer set may  still be defined as an interpretation $J$ that is a least fixpoint of this operator. But, due to non-regularity, some  unexpected answer sets are obtained. 

\begin{example}\label{ex:MR} Take the program:
\begin{gather*}
 s \leftarrow  \SUM(\{1:p, -1:q\}) \geq 0.\\
         q \leftarrow  \SUM(\{1:s\}) > 0. ~~~~~~~
         p \leftarrow  \SUM(\{1:q\}) > 0.
 \end{gather*}
\end{example}

The bodies of these rules are equivalent to $p\lor\neg q$, respectively $s$ and $q$. Therefore, one expects its answer sets to be the same as those of the simplified program:
\begin{gather*}
s \leftarrow  p. ~~~~~~~~ s \leftarrow  \neg q.~~~~~~~~
q \leftarrow  s. ~~~~~~ p \leftarrow  q.
\end{gather*}

To  check that $J=\{p, q, s\}$ is an answer set, we observe that $(\emptyset, J)\models_{\MR} \SUM(\{1:p, -1:q\})\geq 0$, hence the first iteration derives the head $s$. Two more iterations reconstruct the fixpoint $\{p, q, s\}$ which therefore is an answer set according to these semantics.

On the other hand, the Gelfond-Lifschitz reduct of the simplified program with respect to $J = \{p, q, s\}$ is:
\begin{equation*}
s \leftarrow p.~~~~~~
q \leftarrow s. ~~~~~~
p \leftarrow q.
\end{equation*}

Since $\emptyset$ is a model of the reduct, $\{p,q,s\}$ is not an answer set of the simplified program. The culprit for 'too many' answer sets of the aggregate program is the non-regularity of $\models_{MR}$ which leads  to the {\em unsafe}   derivation of aggregate atoms: in the first derivation, $(\emptyset,\{p, q, s\})\models_{MR}\SUM(\{1:p, -1:q\})\geq 0$ which derived $s$, but this derivation was {\em unsafe} since this aggregate atom is not satisfied in the more precise $(\emptyset,\{q\})$. However, things change when looking only at convex aggregate atoms. 

\begin{definition}[Convex aggregate atom]
An aggregate atom $a^{\Aggr}$ is convex iff for all interpretations $I, Z, J$, such that $I \subseteq Z \subseteq J$, it holds that if $I\models_2 a^{\Aggr}$ and $J\models_2 a^{\Aggr}$, then $Z\models_2 a^{\Aggr}$.
\end{definition}

\begin{proposition}
For convex aggregate atoms, $\models_{\MR}$ behaves lower-regular and  equivalent with $\models_{\ult}$.
\end{proposition}

\paragraph{Faber, Pfeifer and Leone (2011).}
They provide semantics for a broader class of ASP programs including negated aggregate atoms. This approach constructs a reduct $P^J$ for a program $P$ with respect to an interpretation $J$ by deleting all rules of which the body is not satisfied in $J$. In general, the immediate consequence operator $T_{P^J}$ is not monotone, so answer sets cannot be defined using the $\lfp$ construction of this operator. Nevertheless, the $\FLP$-answer sets are the answer sets of the following \TSR:

\begin{definition}[$\models_{\FLP}$] We define $\models_{\FLP}$ as follows:
$(I, J)\models_{\FLP} \psi$ iff  $I \models_2 \psi$ and $J \models_2 \psi$.
\end{definition}

\begin{proposition}
$\models_{\FLP}$  extends $\models_2$, i.e., $(I,I) \models_{\FLP} \psi $ iff $I \models_2 \psi$. For conjunctions of aggregate free literals, $\models_{\FLP}$ coincides with $\models_{\GL}$. 
\end{proposition}

This very simple \TSR is neither lower-regular nor lower-monotone, thus the constructive test is inapplicable.

As a consequence, discrepancies between some aggregate programs and their aggregate-free simplification are also present. Using again Example~\ref{ex:MR}, $(\emptyset, \{p, q, s\})\models_{\FLP} \SUM(\{1:p, -1:q\})\geq 0$. But again, $(\emptyset, \{q\})\not\models_{\FLP} \SUM(\{1:p, -1:q\}) \geq 0$. Similarily to $\models_{\MR}$, the $\FLP$-semantics leads to $\{p, q, s\}$ as an answer set. It is obvious that if $(I, J)\models_{\FLP} \psi$, then $(I, J)\models_{\MR} \psi$, since $I \subseteq I$.  Accordingly $\models_{\FLP}$ is less precise then $\models_{\MR}$. Analogously, if $(I, J)\models_{\ult} \psi$, then $(I, J)\models_{\FLP} \psi$ since the interpretations $I$ and $J$ are both elements of $[I, J]$.  

\begin{proposition}
For convex aggregate atoms, the \TSR $\models_{\FLP}$ behaves lower-regular and equivalent to $\models_{\ult}$.
\end{proposition}

\paragraph{\cite{ferraris2011logic}.}
This semantics is closely related to the $\FLP$-semantics. Actually, they  coincide for negation-free programs. \cite{ferraris2011logic} also cover more extensive instances of answer set programs, such as arbitrary propositional theories. It obtains the reduct $P^J$ for a program $P$ and interpretation $J$ by replacing all maximal subformulas in $P$ that are unsatisfied in $J$ by $\bot$. This corresponds to the \TSR $\models_{\F}$:

\begin{definition}[$\models_{\F}$]
$\models_{\F}$ extends $\models_{\GL}$ with: Let $a^{\Aggr} = \Agg(\{a_1:cond_1. \ldots, a_n:cond_n\})*w$, $(I, J) \models_{\F} a^{\Aggr}$ iff $J \models_2 a^{\Aggr}$ and $I \models_2 \Agg(\{a_i:cond_i \in \{a_1:cond_1, \ldots, a_n:cond_n\}|J \models_2 cond_i\})*w$.
\end{definition}

Since $\models_{\F}$ and $\models_{\FLP}$ coincide for negation-free programs, an analogous discussion of Example \ref{ex:MR} leads to the conclusion that the satisfaction relation $\models_{\F}$ is not lower-regular due to the lack of monotonicity. 

\begin{proposition}
$\models_{\F}$  extends $\models_2$, i.e., $(I,I) \models_{\F} \psi $ iff $I \models_2 \psi$.
\end{proposition}

\begin{proposition}
For convex aggregate atoms, $\models_{\F}$ behaves lower-monotone, i.e., if $I \subseteq I'$ and $(I, J) \models_{\F} \psi$, then $(I', J)\models_{\F} \psi$.
\end{proposition}

\begin{proposition}
For anti-monotone aggregate atoms, $\models_{\F}$ behaves lower-regular and equivalent with $\models_{\ult}$.
\end{proposition}

\subsection{Summary}
\begin{table}[H]
    \centering
    \begin{tabular}{l| ccccc}
       & $\LPST$ & $\GZ$\tablefootnote{Note that the $\GZ$-semantics is not defined for programs with conditions under default negation within aggregate-atoms and the table should be interpreted accordingly.} & $\MR$ & $\FLP$ & $\F$\\
       \hline
       Regular & arbitrary & arbitrary & convex & convex & anti-monotone\\
       PDB & ult & triv & ult & ult & ult\\
       $\lfp$ & arbitrary & arbitrary & arbitrary & convex & convex\\
       $<_p$ & $\{\MR, \FLP, \F\}$ & $\{\bnd, \LPST, \MR, $ & $\emptyset$ & $\{\MR\}$ & $\emptyset$ \\
       & & $\FLP, \F\}$ & & \\
    \end{tabular}
    \caption{A summary of the different ternary satisfaction relations.} 
    \label{tab:Summary}
\end{table}
Table~\ref{tab:Summary} gives an overview. Row 1 indicates for which aggregate atoms the ternary satisfaction relation behaves lower-regular. Row 2 shows which semantics from \cite{pelov2007well} coincides with this semantics for the subclass of programs such that the semantics behaves lower-regular. Row 3 indicates for which aggregate atoms the semantics is monotone in the first component such that the answer sets can be constructed as the $\lfp$ of $\ICO{P}$. Row 4 gives the set $S_a$ of semantics discussed in this paper that are strictly more precise than the considered instance $a$. Consequently every $a$-answer will always be an answer set for every semantics in $S_a$. This does not generally hold the other way around. Interestingly, all non-regular semantics coincide with the ult-semantics of \cite{pelov2004semantics} for the subclass of programs in which they behave lower-regular\footnote{Regularity is a property of a semantics. A program cannot be regular or non-regular. However, if a program belongs to a specific subclass of programs, a non-regular semantics may behave lower-regular anyway.}. For programs outside this subclass, the non-regular semantics differ and may derive more answer sets than the ult-semantics.

\section{Conclusions and future work}\label{sec:conclusion}
Approximation Fixpoint Theory describes various types of constructions from  nonmonotonic operators and was designed to formalize  the view of Logic Programming as constructive definitions, a view at least implicit in stratified logic programs,  in the KK and WF semantics, but also in the logic FO(ID) introduced by \cite{denecker2008logic}. We studied aggregate programs from this view point, showing how regular (3- or 4-valued) extensions of the strong Kleene truth assignment induce extensions of KK, WF and stable semantics.  We showed that regular truth assignments correspond one-to-one to  pairs of  a lower-regular and an upper-regular ternary satisfaction relation $(I,J)\models_3 \psi$, where the lower-regular one suffices for defining stable models. To study the relation with ASP, we then made a generalized study of \textit{TSR}s as a tool to define answer sets. We analysed different properties of \textit{TSR}s, and many semantics of aggregate programs in the literature  to determine the corresponding lower ternary satisfaction relation and the properties that influence them,  such as convexity, (anti-)monotonicity, and the sign of conditions in aggregates. We obtained many results linking many ASP semantics in various degrees to the AFT-framework. 

In the  ASP community, other  views of LP and ASP exist than that as a logic of constructive definitions. They are developed in various frameworks such as the framework of \HT (for more details see the paper by \cite{cabalar2017stable}) or of Ferraris and Lifschitz (for more details see the paper by \cite{ferraris2007new}). These frameworks often extend the  original logic programming formalism in various directions, e.g., with disjunction in the head, other negations. They may entirely redefine full FO and the meaning of its connectives. The base idea is that a program corresponds to a theory in some logic (e.g., \HT or FO) from which answer set are derived using some equilibrium characterisation. Although these semantics are not constructive in the sense of AFT, there are surely many interesting mathematical relationships to AFT. For instance, it is striking that the logic of \HT is also defined in terms of a ternary satisfaction relation. It is a goal for future work to investigate this.

\section*{Acknowledgements}
This research received funding from the Flemish Government under the ``Onderzoeksprogramma Artificiële Intelligentie (AI) Vlaanderen'' programme.

\appendix
\setcounter{proposition}{0}
\section{Technical proofs}
\begin{proposition}
$\HH{}^3$ is regular iff $\modelsdown$ is a lower- and $\modelsup$ an upper-regular \TSR. 
\end{proposition}

\begin{proof}
Let $\HH{}^2$ denote the two-valued truth assignment.
\begin{enumerate}
    \item Assume $\HH{}^3$ is regular; we show that $\modelsdown$ is a lower-regular ternary satisfaction relation: 
    \begin{itemize}
        \item $\HH{I}^2(\psi) = v$ with $v \in \{\Fa, \Tr\}$, iff $\HH{(I, I)}^3(\psi) = v$. This entails that $\HH{(I, I)}^3(\psi) = \Tr \textrm{ iff } \HH{I}^2(\psi) = \Tr$. So, $(I,I)\modelsdown \psi$ iff $I\models_2 \psi$ and  $\modelsdown$ extends $\models_2$.
        \item If $(I, J) \leq_p (I', J') \textrm{ then for every }\psi: \HH{(I, J)}^3(\psi) \leq_p \HH{(I', J')}^3(\psi)$. 
        So, if $\HH{(I, J)}^3(\psi)= \Tr =(\Tr, \Tr)$, the truth-value according to $\HH{(I', J')}^3(\psi)$ should also be equal to \Tr. Consequently $(I, J) \modelsdown \psi \textrm{ implies } (I', J') \modelsdown \psi$ (monotone).
    \end{itemize}
    \item Assume $\HH{}^3$ is regular; we show that $\modelsup$ is an upper-regular ternary satisfaction relation: 
    \begin{itemize}
        \item $\HH{I}^2(\psi) = v$ with $v \in \{\Fa, \Tr\}$, iff $\HH{(I, I)}^3(\psi) = v$. This entails that $\HH{(I, I)}^3(\psi) = \Fa \textrm{ iff } \HH{I}^2(\psi) = \Fa$. In other words, $\HH{(I, I)}^3(\psi) \neq \Fa \textrm{ iff } \HH{I}^2(\psi) \neq \Fa$. Based on the derivation of the upper satisfaction relation from the truth assignment, it is clear that $\HH{(I, I)}^3(\psi) \neq \Fa$ iff $(I, I) \modelsup \psi$. On the other hand, for the two-valued truth-assignment, it holds that $\HH{I}^2(\psi) \neq \Fa$ iff $\HH{I}^2(\psi) = \Tr$ iff $I \models_2 \psi$. So, $(I,I)\modelsup \psi$ iff $I\models_2 \psi$ and $\modelsup$ extends $\models_2$.
        \item If $(I, J) \leq_p (I', J') \textrm{ then for every }\psi: \HH{(I, J)}^3(\psi) \leq_p \HH{(I', J')}^3(\psi)$. So, if $\HH{(I, J)}^3(\psi)=\Fa=(\Fa,\Fa)$, the truth-value according to $\HH{(I', J')^3}(\psi)$ should also be equal to \Fa. Consequently $(I, J) \not\modelsup \psi \textrm{ implies } (I', J') \not\modelsup \psi$. Therefore, $(I', J') \modelsup \psi \textrm{ implies } (I, J) \modelsup \psi$ (anti-monotone).
    \end{itemize}
    \item Assume $\modelsdown$ is a lower- and $\modelsup$ an upper-regular ternary satisfaction relation; we show that  $\HH{}^3$ is regular. 
    \begin{itemize}
        \item $(I, I)\modelsdown \psi$ iff $I \models_2 \psi$ iff $(I, I) \modelsup \psi$. Therefore $\HH{(I, I)}^3(\psi) = \Tr$ iff $\HH{I}^2(\psi) = \Tr$ iff $\HH{(I, I)}^3(\psi) \in \{\Tr, \Un\}$. Consequently, $\HH{(I, I)}^3(\psi) = \Fa$ iff $\HH{I}^2(\psi) = \Fa$. Hence $\HH{}^3$ coincides with $\HH{}^2$ for two-valued interpretations.
        \item If $(I, J) \leq_p (I', J')$, then $(I, J) \modelsdown \psi$ implies $(I', J')\modelsdown \psi$. Therefore $\HH{(I, J)}^3(\psi) = \Tr$ implies $\HH{(I', J')}^3(\psi) = \Tr$. At the same time, $(I', J') \modelsup \psi$ implies that $(I, J) \modelsup \psi$. Or equivalently, $(I, J) \not \modelsup \psi$ implies $(I', J') \not \modelsup \psi$. Consequently $\HH{(I, J)}^3(\psi) = \Fa$ implies $\HH{(I', J')}^3(\psi) = \Fa $. If $\HH{(I, J)}^3(\psi) = \Un$ no restrictions are imposed on $\HH{(I', J')}^3(\psi) = \Un$. In all three scenarios it holds that if $(I, J) \leq_p (I', J')$, then $\HH{(I, J)}^3(\psi) \leq_p \HH{(I', J')}^3(\psi)$.
    \end{itemize}
\end{enumerate}
\end{proof}

\begin{proposition}\label{prop:approximator} If  $\modelsdown$ and $\modelsup$ are lower- and upper-regular TSRs, then $A_P$ is an $L^c$ approximator. 
Moreover it is isomorphic to the three-valued $\ICO{P}$ induced by the three-valued truth assignment $\HH{}^3$ combining $\modelsdown$ and $\modelsup$.
\end{proposition}

\begin{proof}
We have to proof that $A_P = (A_P^{\modelsdown}(I, J), A_P^{\modelsup}(I, J)$ is an approximating operator for $\TP{P}$ if $\modelsdown$ and $\modelsup$ are lower- and upper-regular respectively.

\begin{enumerate}
    \item $\leq_p$-monotonicity, i.e., $\forall I, J, I', J': (I, J) \leq_p (I', J') \Leftrightarrow A_P(I, J) \leq_p A_P(I',J')$. $A_P(I, J) \leq_p A_P(I',J')$ is equivalent to $A_P^{\modelsdown}(I, J) \leq A_P^{\modelsdown}(I',J') \wedge A_P^{\modelsup}(I, J) \geq A_P^{\modelsdown}(I',J')$ according to how the precision order is defined. Now, $A_P(I, J) \leq_p A_P(I',J')$ means that, for every atom $p$, it holds that $p \in A_P^{\modelsdown}(I, J) \implies p \in A_P^{\modelsdown}(I',J')$ and $p \in A_P^{\modelsup}(I', J') \implies p \in A_P^{\modelsdown}(I, J)$. If $\modelsup$ is upper-regular, then for every formula $\psi: (I', J') \modelsup \psi \textrm{ implies } (I, J) \modelsup \psi$. So for every rule $ p \leftarrow \psi \in P: (I', J') \modelsup \psi \textrm{ implies } (I, J) \modelsup \psi$. Now, for every atom $p \in A_P^{\modelsup}(I', J')$ there exists a rule $p \leftarrow \psi \in P$, such that $(I', J') \modelsup \psi$ and therefore $(I, J) \modelsup \psi$. Consequently, according to the definition for $A_P^{\modelsup}(I, J)$, it follows that $p \in A_P^{\modelsup}(I, J)$. Analogously, we see that for every atom $p \in A_P^{\modelsdown}(I, J)$ there exists a rule $p \leftarrow \psi \in P$, such that $(I, J) \modelsdown \psi$. If $\modelsdown$ is lower-regular, this entails that $(I', J') \modelsdown \psi$ and therefore $p \in A_P^{\modelsdown}(I', J')$ by definition of $A_P^{\modelsdown}(I', J')$ so $\leq_p$-monotonicity holds.
    \item $A_P$ extends $\TP{P}$, i.e., $\forall I: A_P(I, I) = (\TP{P}(I), \TP{P}(I))$. In other words, $A_P^{\modelsdown}(I, I) = \TP{P}(I))$ and $A_P^{\modelsup}(I, I) = \TP{P}(I))$. $\TP{P}(I))$ is by definition equal to $\{p| p \leftarrow \psi \in P, I \models_2 \psi\}$. Since $\modelsdown$ and $\modelsup$ are lower- and upper-regular respectively, we know that $(I, I)\modelsdown \psi \textrm{ iff } I\models_2 \psi \textrm{ iff } (I, I)\modelsup \psi$. Therefore, $A_P^{\modelsdown}(I, I) = \{p| p \leftarrow \psi \in P, (I, I) \modelsdown \psi \} = \{p|p \leftarrow \psi \in P, I \models_2 \psi\} = \TP{P}(I)) = A_P^{\modelsup}(I, I) = \{p | p \leftarrow \psi \in P, (I, I) \modelsup \psi\}$. \item In conclusion, $A_P(I, J)$ is an approximating operator if $\modelsdown$ and $\modelsup$ are lower- and upper-regular.
    \end{enumerate}
  Finally, we have to prove that $A_P$ is isomorphic to the three-valued $\ICO{P}$. Let $\I = (I, J)$ be a three-valued interpretation.
    \begin{itemize}
        \item Then $A_P(I, J)_1 = A_P^{\modelsdown}(I, J) = \{p | (p \leftarrow \psi) \in P, (I, J)\modelsdown \psi\} = \{p | (p \leftarrow \psi) \in P, \HH{}^3(I, J)(\psi) = \Tr\}$. On the other hand, we know that $\ICO{P}(I, J)_1$ denotes the set of atoms $p$ such that $\lub\{\HH{}^3(I, J)(\psi) | (p \rul \psi) \in P\} = \Tr$ or thus, such that at least one rule with $p$ in the head has a body that evaluates to $\Tr$. This corresponds to the set $\{p | (p \leftarrow \psi) \in P, \HH{}^3(I, J)(\psi) = \Tr\}$. So $A_P(I, J)_1 = \ICO{P}(I, J)_1$.
        \item Then $A_P(I, J)_2 = A_P^{\modelsup}(I, J) = \{p | (p \leftarrow \psi) \in P, (I, J)\modelsup \psi\} = \{p | (p \leftarrow \psi) \in P, \HH{}^3(I, J)(\psi) \in \{ \Tr, \Un\}\}$. $\ICO{P}(I, J)_2$ denotes the set of atoms $p$ such that $\lub\{\HH{}^3(I, J)(\psi) | (p \rul \psi) \in P\} \in \{\Tr, \Un\}$. In words, at least one rule with $p$ in the head should have a body that at least evaluates to $\Un$. This corresponds again to the set $\{p | (p \leftarrow \psi) \in P, \HH{}^3(I, J)(\psi) \in \{ \Tr, \Un\}\}$. So $A_P(I, J)_2 = \ICO{P}(I, J)_2$.
    \end{itemize}
\end{proof}

\begin{proposition}[semi-constructive answer sets]
If $\modelsdown$ is lower-monotone, then for $\LL$-programs $P$, $I$ is an answer set  of $P$ iff $I$ is the  limit of the increasing sequence $\langle I_\alpha\rangle_{\alpha\geq 0}$ where (1) $I_0=\emptyset$, (2) $I_{\alpha+1}=A_P^{\modelsdown}(I_\alpha,I)$ if $I_\alpha\subseteq I$, (3) $I_\lambda = \bigcup_{\alpha<\lambda}I_\alpha$ for limit ordinal $\lambda$. 
\end{proposition}
\begin{proof}
If $\modelsdown$ is lower-monotone, then for $\LL$-programs $P$ we find that $\lambda I: A_P^{\modelsdown}(I, J) = \{p | \exists(p \leftarrow \psi) \in P: (I, J) \modelsdown \psi\}$ is a monotone operator. By the Knaster–Tarski theorem for monotone operators, we know that the described increasing sequence converges to the least fixpoint of the operator. Therefore, all we need to proof is that the least fixpoint of $\lambda J: A_P^{\modelsdown}(J, I)$ is $I$ iff $I$ is an answer set of $P$. $I$ is the least fixpoint of $\lambda J: A_P^{\modelsdown}(J, I)$ iff (i) $A_P^{\modelsdown}(I, I) = I$ and (ii) there is no $J \subset I$ such that $A_P^{\modelsdown}(J, I) = J$. 
\begin{enumerate}
\item (i) and (ii) $\implies$ $I$ is an answer set. \\
If $A_P^{\modelsdown}(J, I) = J$, then for every rule $p \leftarrow \psi \in P$, it holds that if $(J, I)\modelsdown \psi$, then $p \in J$ and thus $(J, I) \modelsdown p$. Thus if (i), then for every $(p\leftarrow\psi)\in P$, if $(I, I) \modelsdown\psi$ and hence $I\models_2\psi$, then $(I, I)\modelsdown p$ and hence $I\models_2 p$. This is condition (1) of the answer set definition (Definition~\ref{def:GAS}). 

Since $\lambda J: A_P^{\modelsdown}(J, I)$ is monotone and $I$ is the least fixpoint, we find that from (ii), it follows that for every $J \subset I$, there exists a rule $(p\leftarrow\psi)\in P$, such that  $(J, I)\modelsdown \psi$ and $(J, I)\not \modelsdown p$, i.e., there is no $J \subset I$ such that for every rule $(p\leftarrow\psi)\in P$, if  $(J, I)\modelsdown \psi$ then $(J, I) \modelsdown p$. This is condition (2) of Definition~\ref{def:GAS}, so $I$ is an answer set.

\item $I$ is an answer set $\implies$ (i).\\
If $I$ is an answer set, then for every $(p\leftarrow\psi)\in P$, if $I \models_2 \psi$, then $I\models_2 p$. By definition of $A_P^{\modelsdown}$, this entails that $A_P^{\modelsdown}(I, I)= I' \subseteq I$. Assume $I' \subset I$, then for every rule $p \leftarrow \psi \in P$, if $(I', I)\modelsdown \psi$, then by lower-monotonicity of $\modelsdown$, $(I, I)\modelsdown \psi$, hence by definition of $A_P^{\modelsdown}$, $p \in A_P^{\modelsdown}(I, I)=I'$, which entails $(I', I) \modelsdown p$. Therefore $I'$ violates condition (2) of the answer set  definition.
\item $I$ is an answer set $\implies$ (ii).\\
Assume that there exists a $J \subset I$, such that $A_P^{\modelsdown}(J, I)= J$, then it holds that for every rule $(p\leftarrow\psi)\in P$, if $(J, I)\modelsdown \psi$, then $(J, I) \modelsdown p$, so $J$ violates condition (2) of the answer set definition.
\end{enumerate}
\end{proof}

\begin{proposition}\label{prop:AFTstable} 
Answer sets and AFT-stable models coincide for $\LL$ programs with a lower regular TSR.
\end{proposition}

\begin{proof}
If $\modelsdown$ is lower-regular, then the AFT-stable models are defined. Recall from Section \ref{sec:AFT} that, according to the constructive test, $x$ is a stable model iff $x$ is the limit of the $\lfp$ construction of $\lambda z: A(z,x)_1$ with $A(z, x)$ an approximator on $L^c$. From Proposition \ref{prop:approximator}, we know that $A_P(I,J) = (A_P^{\modelsdown}(I, J), A_P^{\modelsup}(I, J))$ is an approximator for $T_P$ if $\modelsdown$ and $\modelsup$ are lower- and upper-regular. This holds as long as we choose an upper-regular $\modelsup$. Then the first component $A_P(I,J)_1$ corresponds to $A_P^{\modelsdown}(I, J)$. Thus $J$ is a stable model iff $J$ is the limit of the $\lfp$ construction of $\lambda I: A_P^{\modelsdown}(I, J)$ iff $J$ is an answer set (by Proposition \ref{prop:semi-constructive}).
\end{proof}

\begin{proposition}
For $I\in[\bot, J]$, $I\models_2 P^J$  iff  for every rule $p\leftarrow\psi\in P$, if $(I,J)\models_{GL} \psi$ then $(I,J)\models_{GL} p$. \ignore{If $P$ is non-disjunctive, then}$J$ is a GL-answer set of $P$ iff $J$ is an AFT-stable model of $P$ under strong Kleene truth assignment. 
\end{proposition}

\begin{proof}
\begin{enumerate}
\item For every rule $r^J$ in $P^J$ there exists a counterpart $r$ in $P$, however, the opposite is not true. Since the head of a rule is never changed in the reduct, proving equivalence of the relations corresponds to showing for every rule $p \leftarrow \psi$ that $(I, J)\models_{\GL} \psi$ if and only if $(p \leftarrow \psi)^J$ exists and $I \models_2 \psi^J$. Any rule $p \leftarrow \psi$ in $P$ is of the form $p \leftarrow l_1 \wedge ... \wedge l_n \wedge \neg t_1 \wedge ... \wedge \neg t_m$ with $l_1, ..., l_n, t_1, ..., t_m$ atoms. If there exists an atom $t_i$ such that $t_i \in J$, then the rule is deleted during the first step of the construction of the reduct. Consequently, for every $I$ we expect $(I, J)\not\models_{GL} \psi$. Indeed, from $t_i \in J$ the satisfaction relation deduces that $(I, J) \not\models_{GL} \neg t_i$ and therefore $(I, J) \not\models_{GL} \psi$. If for every $j \in [1, m]$ it holds that $t_j \notin J$, then $p \leftarrow \psi$ is transformed to $(p \leftarrow \psi)^I$ during the second step of the reduct-construction. This new rule is now given by $p \leftarrow l_1 \wedge ... \wedge l_n$. From the definition of the satisfaction relation $\models_2$ we know that $I\models_2 \psi^J$ if and only if $l_1,..., l_n \in I$ and because $(I, J) \models_{GL} \neg t_i$ for all $i$, we have $(I, J)\models_{GL} \psi$. Hence $I\models_2 \psi^J$ iff $(I, J)\models_{GL} \psi$.
\item We now know that for $I\in[\bot, J]$, $I\models_2 P^J$  iff $(I,J)\models_{GL} P$ with $P^J$ the GL-reduct of $P$ for $J$. Since $\models_{GL}$ is lower-regular and $P$ is non-disjunctive, we know that $J$ is an AFT stable model iff $J$ is an answer set of $P$ by Proposition \ref{prop:AFTstable}. $J$ is an answer set of $P$ iff $J \models_2 P$ and there is no $I \subset J$ such that $(I, J)\models_{GL} P$. Hence, $J$ is an answer set of $P$ iff $J\models_2 P$ and there is no $I \subset J$ such that $I\models_2 P^J$ iff $J$ is a GL-answer set.
\end{enumerate}
\end{proof}

\begin{proposition}
Let $\models_a, \models_b$ be TSR's that coincide with $\models_{GL}$ on aggregate free bodies. If $\models_a \leq_p \models_b$ and $J$ is an answer set associated with $\models_a$ (an $a$-answer set), then $J$ is a $b$-answer set.
\end{proposition}

\begin{proof}
Assume that $J$ is an $a$-answer set. This means that (i) for every $(p \leftarrow \psi)\in P$, if $J \models_2 \psi$, then $J \models_2 p$ and (ii) for every $I \subset J$ it holds that there exists a rule $(p\leftarrow\psi)\in P$ such that $(I, J) \models_a \psi$ and $(I, J) \not\models_a p$. For $J$ to be a $b$-answer set, observe that (i) holds as it is a $a$-answer set; it remains to show that (ii) holds. Consider the same rule as above. By $\models_a \leq_p \models_b$, we find that $(I, J) \models_b \psi$. Moreover, in the considered programs, $p$ can only be a propositional atom. If $(I, J) \not \models_a p$, then $p \not \in I$ and thus $(I, J) \not \models_b p$. Consequently, $J$ is a $b$-answer set of $P$. 
\end{proof}

\begin{proposition}
The TSRs $\models_{\triv}$, $\models_{\ult}$ and $\models_{\bnd}$ are lower-regular. Since $\models_{triv}\leq_p \models_{bnd} \leq_p \models_{\ult}$, an answer set of $\models_{triv}$ is one of $\models_{bnd}$, and one of $\models_{bnd}$   is one of $\models_{\ult}$.
\end{proposition}

\begin{proof}
From definition 7.1 in \cite{pelov2007well}, it follows that $\HH{}^{triv}, \HH{}^{ult}$ and $\HH{}^{bnd}$ are regular. Then the first part of this proposition follows from Proposition \ref{prop:derived3truth}, the second part from Proposition~\ref{prop:precisionmonotone}.
\end{proof}

\begin{proposition}
For aggregate programs containing only positive conditions in aggregate atoms,  the TSR 
$\models_{\GZ}$  is identical to the TSR $\models_{\triv}$ and lower-regular for consistent pairs, i.e., with $(I, J)$ a consistent pair,
$(I, J) \models_{\GZ} a^{\Aggr} $ iff $(I, J) \models_{\triv} a^{\Aggr}$.
\end{proposition}
    
\begin{proof}  
We only need to proof this for aggregate atoms as $\models_{GZ}$ is truth functional.
\begin{enumerate}
\item $(I, J)\models_{\GZ} a^{\Aggr}$ entails $(I, J)\models_\triv a^{\Aggr}$. By definition of $\models_{\GZ}$, $(I, J) \models_{\GZ} a^{\Aggr}$ iff $J\models_2 a^{\Aggr}$ and $(I, J) \models_{\GZ} \bigwedge \{cond_j \in Cond(a^{\Aggr})\footnote{For $a^{\Aggr} = \Agg(\{a_1: cond_1, \cdot, a_n:cond_n\})*w$, $Cond(a^{\Aggr}) = \{cond_i| i \in [1, n]\}$}| J \models_2 cond_j\}$. The conditions are positive, so they are evaluated in $I$, hence, given that $(I, J) \models_{\GZ} a^{\Aggr}$, $(I, J) \models_{\GZ} \bigwedge \{cond_j \in Cond(a^{\Aggr})| J \models_2 cond_j\}$ and $I \models_2 \bigwedge \{cond_j \in Cond(a^{\Aggr})| J \models_2 cond_j\}$ and, for conditions in this set, $cond_i^I = cond_i^J$. For the other conditions (such that $J \not \models_2 cond_i$), they are false in $J$ hence also false in $I \subseteq J$. So, for all conditions, $cond_i^I = cond_i^J$ and $(I, J)\models_\triv a^{\Aggr}$.

\item $(I, J)\models_{\triv} a^\Aggr$ entails $(I, J)\models_\GZ a^\Aggr$.\\
By definition of $\models_\triv$, if $(I, J) \models_\triv a^\Aggr$, then $J\models_2 a^\Aggr$ and ${cond_i}^J = {cond_i}^I$ for every condition $cond_i \in Cond(a^\Aggr)$. Therefore, for every condition $cond_i \in \{cond_j \in Cond(a^{\Aggr})| J \models_2 cond_j\}$ it holds that $I \models_2 cond_i$. By definition of $\models_{\GZ}$ for conjunction of literals, then $(I, J) \models_{\GZ} \bigwedge \{cond_j \in Cond(a^{\Aggr})| J \models_2 cond_j\}$. Hence, $(I, J) \models_{\GZ} a^\Aggr$.

\item Since the satisfaction relation $\models_{\GZ}$ is equivalent to $\models_{\triv}$ for negation-free aggregate atoms and $\models_{\triv}$ is lower-regular, $\models_{\GZ}$ is also lower-regular.
\end{enumerate}
\end{proof}

\begin{proposition}
(i) $\models_{\MR}$  extends $\models_2$, i.e., $(I,I) \models_{\MR} \psi $ iff $I \models_2 \psi$. 
(ii) $\models_{\MR}$ is 
lower-monotone, i.e., if $I \subseteq I'$ and $(I, J) \models_{\MR} \psi$, then $(I', J)\models_{\MR} \psi$. 
\end{proposition}

\begin{proof}
\begin{enumerate}
    \item The satisfaction relation is truth-functional hence we only need to prove this for aggregate atoms. We know that $(I, I) \models_{\MR} a^{\Aggr}$ iff  $I\models_2 a^{\Aggr}$ and there exists an interpretation $Z \subseteq I$ such that $Z\models_2 a^{\Aggr}$. Take $Z = I$ and it is clear that $I\models_2 a^{Aggr}$ implies that there exists a $Z$ that satisfies the requirements. This observation leads to $(I, I) \models_{\MR} a^{\Aggr}$ iff  $I\models_2 a^{\Aggr}$.
    \item The part of the definition for $\models_{\MR}$ that was taken from $\models_{\GL}$ satisfies this property since $(I, J) \leq_p (I', J)$ and $\models_{\GL}$ is lower-regular. For the rule regarding aggregates, it is clear that if $(I, J) \models_{\MR} a^{\Aggr}$, then $J \models_2 a^{\Aggr}$ and there exists a $Z \subseteq I \subseteq I'$ such that $Z \models_2 a^{\Aggr}$. Clearly, this means that $(I', J)\models_{\MR} a^{\Aggr}$, so $\models_{MR}$ is lower-monotone.
\end{enumerate}
\end{proof}

\begin{proposition}\label{prop:MRconvex}
For  convex aggregate atoms,  $\models_{\MR}$ behaves lower-regular and equivalent with $\models_{\ult}$.
\end{proposition}
\begin{proof}
\begin{enumerate}
\item We have already shown that it extends the satisfaction relation $\models_2$ for arbitrary aggregate atoms, including convex ones. All that is left to prove is its $\leq_p$-monotone behaviour. If $(I, J) \leq_p (I', J')$ and $(I, J) \models_{\MR} a^{\Aggr}$ with $a^{\Aggr}$ a convex aggregate atom, then $J \models_2 a^{\Aggr}$ and there exists an interpretation $Z \subseteq I$ such that $Z\models_2 a^{\Aggr}$. According to the definition of convex aggregate atoms, this implies that for every interpretation $Z'$ such that $Z \subseteq Z' \subseteq J$, $Z' \models_2 a^{\Aggr}$. This includes the interpretations $I, I'$ and $J'$. From this, it is trivial to see that $(I', J') \models_{\MR} a^{\Aggr}$. 
\item If $(I, J) \models_{\MR} a^{\Aggr}$, then $J \models_2 a^{\Aggr}$ and there exists an interpretation $Z \subseteq I$ such that $Z\models_2 a^{\Aggr}$. Since $a^{\Aggr}$ is convex, this means that for every $X$ such that $Z \subseteq (I \subseteq) X \subseteq J$, $X\models_2 a^{\Aggr}$. Thus, $(I, J) \models_{\ult} a^{\Aggr}$. 
\item If $(I, J) \models_{\ult} a^{\Aggr}$, then $X\models_2 a^{\Aggr}$ for every $I \subseteq X \subseteq J$, thus $J \models_2 a^{\Aggr}$ and $I \models_2 a^{\Aggr}$ with $I \subseteq I$. Therefore, $(I, J) \models_{\MR} a^{\Aggr}$.
\end{enumerate}
\end{proof}

\begin{proposition}
$\models_{\FLP}$  extends $\models_2$, i.e., $(I,I) \models_{\FLP} \psi $ iff $I \models_2 \psi$. For conjunctions of aggregate free literals, $\models_{\FLP}$ coincides with $\models_{\GL}$. 
\end{proposition}

\begin{proof}
\begin{enumerate}
    \item $\models_\FLP$ extends $\models_2$:\\
    By definition $(I, I)\models_\FLP \psi$ iff $I \models_2 \psi$ and $I \models_2 \psi$. Hence $(I, I)\models_\FLP \psi$ iff $I \models_2 \psi$.
    \item $\models_\FLP$ coincides with $\models_{\GL}$ for conjunctions of aggregate free literals:\\
    For literals $l_i$ it is easy to see that: $(I, J) \models_\FLP l_i$ iff $(I, J) \models_\GL l_i$ since if $I \models_2 p$, then $J \models_2 p$ and if $J \models_2 \neg p$, then $I \models_2 \neg p$. For conjunctions of literals $\bigwedge_{i=1}^{n} l_i$, we have $(I, J) \models_\FLP \bigwedge_{i=1}^{n} l_i$ iff $I \models_2 \bigwedge_{i=1}^{n} l_i$ and $J \models_2 \bigwedge_{i=1}^{n} l_i$. By the definition of $\models_2$ for conjunctions, this is equivalent to $I \models_2 l_i$ and $J \models_2 l_i$ for every $l_i$ in the conjunction. This in turn is equivalent to $(I, J) \models_\FLP l_i$. By means of the aforementioned result for literals, this is equivalent with $(I, J) \models_\GL l_i$. By definition of $\models_\GL$ for conjunctions of literals, we then find $(I, J) \models_\FLP \bigwedge_{i=1}^{n} l_i$ iff  $(I, J) \models_\GL \bigwedge_{i=1}^{n} l_i$.
\end{enumerate}
\end{proof}

\begin{proposition}
 For convex aggregate atoms, the ternary satisfaction relation $\models_{\FLP}$ behaves lower-regular and equivalent to $\models_{\ult}$.
\end{proposition}
\begin{proof}
\begin{enumerate}
    \item Firstly, we proof: If $(I, J) \models_{\ult} a^\Aggr$, then $(I, J) \models_{\FLP} a^\Aggr$.\\
     If $(I, J) \models_{\ult} a^{\Aggr}$, then $X\models_2 a^{\Aggr}$ for every $I \subseteq X \subseteq J$. Consequently, $I \models_2 a^{\Aggr}$ and $J\models_2 a^{\Aggr}$, hence $(I, J) \models_{\FLP} a^{\Aggr}$.
     \item Secondly, we proof: If $(I, J) \models_{\FLP} a^{\Aggr}$, then $(I, J) \models_{\MR} a^{\Aggr}$.\\
     If $(I, J) \models_{\FLP} a^{\Aggr}$, then $I\models_2 a^{\Aggr}$ and $J\models_2 a^{\Aggr}$. Thus $J \models_2 a^{\Aggr}$ and $I \models_2 a^{\Aggr}$ with $I \subseteq I$. Therefore, $(I, J) \models_{\MR} a^{\Aggr}$.
     \item This means that $\models_{\FLP}$ is placed between $\models_{\ult}$ and $\models_{\MR}$ in the precision order. However, for programs with only convex aggregates, $\models_{\ult}$ and $\models_{\MR}$ coincide. Therefore, $\models_{\FLP}$ will also coincide with these semantics.
     \item Consequently, $\models_{\FLP}$ is a lower-regular relation when we only consider convex aggregates.
\end{enumerate} 
\end{proof}

\begin{proposition}\label{prop:FExtends}
$\models_{\F}$  extends $\models_2$, i.e., $(I,I) \models_{\F} \psi $ if $I \models_2 \psi$.
\end{proposition}
\begin{proof}
The truth-assignment is truth-functional, therefore, we only need to prove the property for aggregate atoms. We know that $(I, I) \models_{\F} a^{\Aggr}$ iff $I \models_2 a^{\Aggr}$ and $I \models_2 \Agg (\{a_i : cond_i \in \{a_1: cond_1, \ldots, a_n : cond_n\}|I \models_2 cond_i \}) * w$. It follows that if $I \models_2 a^{\Aggr}$, then $I \models_2 \Agg (\{a_i : cond_i \in \{a_1: cond_1, \ldots, a_n : cond_n\}|I \models_2 cond_i \}) * w$. This observation leads to  $(I, I) \models_{\F} a^{\Aggr}$ iff $I \models_2 a^{\Aggr}$.
\end{proof}
  
\begin{proposition}\label{prop:Ffirstmonotone}
For convex aggregate atoms, $\models_{\F}$ behaves lower-monotone, i.e., if $I \subseteq I'$ and $(I, J) \models_{\F} \psi$, then $(I', J)\models_{\F} \psi$.
\end{proposition}  
\begin{proof}
$\models_{\F}$ extends $\models_{\GL}$ and the latter is lower-monotone. So we only need to consider a $\psi$ that is a convex aggregate atom. If $a^\Aggr$ is a convex aggregate atom, then $a^{\Aggr}_J = \Agg (\{a_i : cond_i \in \{a_1: cond_1, \ldots, a_n : cond_n\}|I \models_2 cond_i \}) * w$ is also convex since the transformation only deletes conditions from the original convex aggregate. This means that if  $X \subseteq Y \subseteq Z$, $X \models_2 a^{\Aggr}_J$, and $Z \models_2 a^{\Aggr}_J$, then $Y \models_2 a^{\Aggr}$. It is easy to see that if $J \models_2 a^{\Aggr}$, then $J \models_2 a^{\Aggr}_J$. Therefore, if $(I, J)\models_{\F} a^{\Aggr}$ and $(I, J) \leq_p (I', J)$, then $I' \models_2 a^{\Aggr}_J$ and thus $(I', J) \models_{\F} a^{\Aggr}$.
\end{proof}

\begin{proposition}
For anti-monotone aggregate atoms, $\models_{\F}$ behaves lower-regular and equivalent with $\models_{\ult}$.
\end{proposition}

\begin{proof}
Let $a^\Aggr_J = \Agg (\{a_i : cond_i \in \{a_1: cond_1, \ldots, a_n : cond_n\}|J \models_2 cond_i \}) * w$.
\begin{enumerate}
    \item If $(I, J) \models_{\ult} a^\Aggr$, then $X\models_2 a^\Aggr$ for every $I \subseteq X \subseteq J$. Thus, $J \models a^\Aggr$ and therefore $J \models a^\Aggr_J$. Since $a^\Aggr$ is anti-monotone and the only transformation between $a^\Aggr$ and $a^\Aggr_J$ deletes conditions, $a^\Aggr_J$ must be anti-monotone. Hence $I \models a^\Aggr_J$ and $(I, J) \models_{\F} a^\Aggr$.
    \item If $(I, J) \models_{\F} a^{Aggr}$, then $J \models_2 a^\Aggr$. From the anti-monotonicity of $a^\Aggr$ it follows that for all $X \subseteq J$, $X \models_2 a^\Aggr$. Hence, $(I, J) \models_{\ult} a^\Aggr$.
    \item An anti-monotone aggregate atom is convex and $\models_{\F}$ and $\models_{\ult}$ coincide for convex aggregate atoms. Therefore, for anti-monotone aggregate atoms, $\models_{\F}$ behaves lower-regular.
\end{enumerate}
\end{proof}

\end{document}